\documentclass{article}

\usepackage{arxiv}

\usepackage[utf8]{inputenc} 
\usepackage[T1]{fontenc}    
\usepackage{hyperref}       
\usepackage{url}            
\usepackage{booktabs}       
\usepackage{amsfonts}       
\usepackage{nicefrac}       
\usepackage{microtype}      
\usepackage{lipsum}		
\usepackage{graphicx}
\usepackage{natbib}
\usepackage{doi}

\usepackage{microtype}
\usepackage{graphicx}
\usepackage{subcaption}
\usepackage{booktabs} 

\usepackage{hyperref}

\usepackage{amsmath}
\usepackage{amssymb}
\usepackage{mathtools}
\usepackage{amsthm}

\usepackage[capitalize,noabbrev]{cleveref}
\usepackage{enumitem}

\theoremstyle{plain}

\theoremstyle{definition}

\theoremstyle{remark}

\usepackage[textsize=tiny]{todonotes}

\title{On the Role of Inductive Bias in Time-Series Pretraining: A Case Study in Learning Generalizable Representations for Clinical Time Series}

\author{ \href{https://orcid.org/0000-0001-8058-4867}{\includegraphics[scale=0.06]{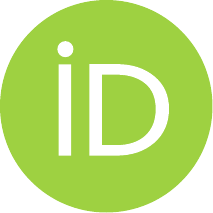}\hspace{1mm}Sharmita Dey}\thanks{Corresponding author} \\
	ETH Zurich\\
    Zurich, Switzerland \\
	\texttt{contact.deysharmita@gmail.com} \\
	\And
	Diego Paez-Granados \\
	ETH Zurich, \\
    Swiss Paraplegic Research, Nottwil \\
    Zurich, Switzerland
}

\renewcommand{\shorttitle}{Role of Inductive Bias in Time-Series Pretraining}

\hypersetup{
pdftitle={Role of Inductive Bias in Time-Series Pretraining},
pdfsubject={q-bio.NC, q-bio.QM},
pdfauthor={Sharmita Dey},
pdfkeywords={First keyword, Second keyword, More},
}

\begin{document}
\maketitle

\begin{abstract}
Clinical time-series learning is routinely constrained by small, heterogeneous cohorts and protocol drift, while its downstream use spans both classification (e.g., pathology diagnosis) and regression (e.g., temporal forecasting). These constraints make foundation-model pretraining appealing, but raises an important question of \textit{which inductive biases should the pretraining objective impose so that representations transfer across task types  and subjects}. We study this question in pathological gait analysis for spinal cord injury (SCI) via \textsc{PathoFM}, an encoder-centric transformer pretrained on multivariate gait windows with three complementary objectives: Local Completion (reconstruct contiguous masked spans to enforce local structure), Temporal Continuity (predict a masked mid-horizon continuation from an observed prefix to enforce smoothness and causal consistency), and Unsupervised In-Context Dynamics (support-query reconstruction conditioned on subject exemplar windows via attention). Empirically comparing objective families (grouping/contrastive, dynamics-based, and generative reconstruction), we find that dynamics-centric mixtures produce the most balanced transfer: grouping objectives favor discriminative margins but can degrade magnitude fidelity needed for continuous targets, whereas reconstruction-only objectives preserve waveform structure but may underperform on classification. Overall, combining local reconstruction with temporal continuity, and adding in-context conditioning when exemplar access is realistic, yields robust subject-generalizing representations. 
\end{abstract}
\vspace{0.2em}

\section{Introduction}
Clinical time series, such as pathological kinematics and kinetics, encode rich information about physiological state, pathology, and progression \cite{sutherland2005evolution}, but they also embody the peculiar constraints of medicine: cohorts are small, labels are expensive, and data are collected under heterogeneous protocols (different clinics, sensors, patient populations) \cite{zantvoort2024estimation, guan2021domain}. These constraints create two recurring failure modes: (i) supervised models overfit cohort idiosyncrasies and generalize poorly to new subjects \cite{zantvoort2024estimation}, and (ii) representations optimized for one endpoint (say, diagnosis) transfer poorly to another (say, a continuous kinetic target) \cite{zhang2022survey, harutyunyan2019multitask}.

Foundation-model pretraining offers a partial escape hatch \cite{brown2020language, bommasani2021opportunities}: by learning representations from large pools of unlabeled sequences, we can amortize feature learning and reduce the label burden on downstream tasks \cite{hinton2006reducing, erhan2010does}. Yet in clinical time series, pretraining is not only an optimization trick; it is a way of committing to an \emph{inductive bias}, a set of assumptions about what structure matters in the data \cite{mitchell1980need}. Contrastive objectives \cite{chen2020simclr}, for example, impose a ``grouping'' bias (instances should be separable and invariant to nuisances), while forecasting objectives impose a ``dynamics'' bias (future should be predictable from past and trajectories should be temporally consistent) \cite{oord2018representation, li2023ti}. The central question we address is:

\emph{
\textbf{Which inductive biases matter for clinical time-series pretraining when the goal is cross-task transfer to both classification and regression under subject shift?}
}

\paragraph{Case study: pathological gait for SCI.}
We ground our study in gait analysis for spinal cord injury (SCI), where each trial yields multivariate gait cycles (kinematics and kinetics) and downstream endpoints include both categorical labels (e.g., paraplegic vs. tetraplegic patterns, functional independence scores) and continuous targets (e.g., ground reaction force components). Gait is a particularly sharp microscope for inductive bias: clinically relevant cues can be local and phase-specific (e.g., swing-phase anomalies), while clinically useful kinetics depend on faithful modeling of magnitudes, phase, and temporal dynamics \cite{perry2024gait, harris2022gaitsurvey, winter2009biomechanics}.

\paragraph{Approach.}
We develop \textsc{PathoFM}, an encoder-only transformer pretrained on gait windows with three complementary objectives: (i) \textit{Local Completion (LC)}: reconstruct contiguous masked spans (a masked-autoencoding bias), (ii) \textit{Temporal Continuity (TC)}: predict masked mid-horizon futures from observed prefixes (a dynamics bias), and (iii) \textit{Unsupervised In-Context Dynamics (uICD)}: reconstruct masked query windows while attending to same-subject support windows (an in-context adaptation bias). The implementation accompanying this paper (summarized in Sec.~\ref{sec:impl}) further includes interpolation-based missing-value handling and a subject-balanced batch sampler to mitigate subject imbalance during pretraining.

\paragraph{Contributions.}
This paper makes the following contributions:
\begin{enumerate}
  \item We present a practical taxonomy of inductive biases for time-series pretraining (Sec.~\ref{sec:taxonomy}) and discuss favored transfers and failure modes. 
  \item We formalize \textsc{PathoFM}'s three-objective pretraining as a dynamics-centric mixture with an explicit in-context component (Sec.~\ref{sec:method}).
  \item We compare objectives spanning grouping, dynamics, and generative families, showing empirically that a dynamics-centric mixture yields balanced transfer across classification and regression under strict subject holdout (Sec.~\ref{sec:results}). 
\end{enumerate}

\section{Related Work and Context}
Self-supervised learning has become a standard approach to representation learning in domains where labels are scarce but unlabeled data are plentiful \cite{gui2024survey}. In clinical time series, however, the question is not merely whether self-supervision helps, but \emph{which self-supervised biases align with clinically relevant endpoints}.

\paragraph{Grouping-based objectives.}
Contrastive learning \cite{le2020contrastive} and prototype-based clustering \cite{li2020prototypical} impose a bias that nearby augmentations of an instance should map to nearby embeddings, while different instances should be separated. This is instantiated by InfoNCE-style objectives (popularized in vision by SimCLR~\cite{chen2020simclr} and MoCo \cite{he2020momentum}) and by prototype methods such as DINO~\cite{caron2021dino}. In biomedical time series, similar ideas are used by treating each subject as a ``class'' (subject-ID pretraining) or by contrasting different windows from the same recording \cite{liu2023self, ogg2024self, yue2022ts2vec}. These objectives often yield strong classification features but can suppress absolute magnitude information and fine-grained waveform details when augmentations encourage invariance to amplitude or timing \cite{wen2020time}.

\paragraph{Masked reconstruction.}
Masked modeling learns representations by reconstructing missing input tokens; it is well known in language (masked language modeling) and has become a standard in vision via masked autoencoders (MAE)~\cite{he2022mae}. In time series, masked span reconstruction tends to preserve local signal morphology and cross-channel correlations \cite{li2023ti}. For clinical signals, this is attractive because many downstream tasks, from denoising to biomechanical interpretation, require preserving clinically meaningful waveform structure, not only class separability.

\paragraph{Forecasting and dynamics learning.}
Forecasting objectives (next-step prediction, multi-horizon prediction, or continuation) explicitly pressure representations to encode temporal dynamics \cite{oord2018representation}. In forecasting-centric architectures such as Temporal Fusion Transformers~\cite{lim2021tft}, the inductive bias is that future outcomes depend predictably on past context \cite{dey2024continual}. For representation learning, forecasting can be used as a pretext to learn phase-consistent latent dynamics even when the eventual downstream task is not forecasting \cite{zhang2024self}. A known concern is exposure bias: a model trained on short horizons may learn brittle dynamics that do not transfer \cite{bengio2015scheduled, venkatraman2015improving}.

\paragraph{Diffusion and probabilistic reconstruction.}
Diffusion models for time series \cite{tashiro2021csdi, dey2025cross} replace point reconstruction with distribution modeling, capturing uncertainty and rich local structure. This can be beneficial for continuous-valued targets and missing-data regimes, but diffusion training typically incurs higher compute and may not optimize for discriminative margins \cite{ho2020denoising, nichol2021improved, skiers2025joint}.

\paragraph{In-context adaptation and meta-learning.}
In-context learning refers to models that adjust behavior at inference time based on a prompt or support set, without parameter updates. In time series, this can be emulated by presenting multiple windows in a support--query structure and training the model to solve a masked reconstruction conditioned on supports \cite{das2024context, lu2024context}. This resembles non-parametric meta-learning and, in the clinical setting, mirrors the common practice of comparing a patient's measurements to reference exemplars \cite{vinyals2016matching, schmidt2001cased}.

\paragraph{Why gait is a good stress test.}
Gait analysis sits at the intersection of discriminative and generative needs: clinicians may want to classify pathology type, but also to estimate continuous kinetic variables and understand phase-specific deviations \cite{dey2024remap, dey2022function, dey2021hybrid, dey2019support, dey2020continuous, dey2019random, quintero2018continuous}. A representation that is ``good for classification'' but poor at magnitude fidelity is therefore incomplete. Standard gait references provide biomechanical context and typical waveform structure~\cite{perry2024gait}, while machine learning surveys emphasize the diversity of gait-based tasks~\cite{harris2022gaitsurvey}. This makes gait a useful case study for objective-induced inductive bias.

\section{A Taxonomy of Inductive Bias in Time-Series Pretraining}
\label{sec:taxonomy}
Pretraining objectives can be understood as imposing preferences about which information should be retained or discarded by the representation. We group common objectives into four families by their dominant inductive bias (Table \ref{tab:taxonomy}). The families are not mutually exclusive, but they predict downstream behavior surprisingly well.

\begin{table*}[t]
\centering
\caption{Taxonomy of inductive biases for time-series pretraining. ``Favored transfers'' are task types that typically benefit most; ``failure modes'' are common when a bias is over-applied.}
\label{tab:taxonomy}
\begin{tabular}{p{0.12\linewidth} p{0.35\linewidth} p{0.22\linewidth} p{0.24\linewidth}}
\toprule
\textbf{Inductive bias} & \textbf{Representative objectives} & \textbf{Favored transfers} & \textbf{Common failure modes} \\
\midrule
\textbf{Grouping in latent space} 
& Contrastive instance discrimination (InfoNCE \cite{oord2018representation}); prototype clustering (e.g., DINO \cite{caron2021dino}); supervised subject-ID \cite{ogg2024self}
& Category classification, cohort stratification, retrieval
& Loses fine-scale waveform fidelity and calibrated magnitude; can over-emphasize between-subject differences \\
\addlinespace
\textbf{Generative reconstruction}
& Masked span modeling (MAE-style) \cite{he2022mae, li2023ti}; conditional diffusion for imputation \cite{tashiro2021csdi}; denoising \cite{ho2020denoising, nichol2021improved}
& Denoising, imputation, signal restoration; features with high SNR
& May under-emphasize discriminative margins; training is costly (e.g., diffusion) \\
\addlinespace
\textbf{Dynamics learning}
& Next/mid-horizon forecasting \cite{lim2021tft}; continuation; autoregressive modeling \cite{bengio2015scheduled, venkatraman2015improving}. 
& Trajectory prediction; continuous clinical endpoints sensitive to timing/rate-of-change
& Exposure bias; overfit to short-term trends; may blur class boundaries if dynamics dominate \\
\addlinespace
\textbf{In-context adaptation}
& Support--query conditioning \cite{dong2024survey}; non-parametric meta-learning \cite{vinyals2016matching, allen2018variadic}. 
& Personalization; cross-subject transfer under distribution shift
& Data-inefficient if supports are poor; possible memorization of idiosyncrasies \\
\bottomrule
\end{tabular}
\end{table*}
\vspace{-0.5cm}
\paragraph{Why this taxonomy matters.}
Clinical workloads rarely involve a single endpoint \cite{harutyunyan2019multitask}. In gait, classification tasks such as diagnosis \cite{slijepcevic2021explaining, alaqtash2011automatic} and decision making \cite{althoff1998case} prefer representations that emphasize discriminative margins (grouping), while regression tasks such as kinetics prediction \cite{liu2022deep, liu2015clinical} or clinical forecasting prefer representations that preserve magnitudes and phase-resolved dynamics (dynamics + reconstruction). A foundation model intended for \emph{both} must therefore balance biases, rather than optimizing any one to extremality \cite{sener2018multi}.

\section{\textsc{PathoFM}: Multi-Objective Pretraining with Complementary Biases}
\label{sec:method}
\subsection{Problem setup and notation}
Let $\mathbf{X} \in \mathbb{R}^{T \times D}$ denote a multivariate gait window of length $T$ with $D$ features. In the reference implementation, we use fixed-length windows with a \emph{past} portion and a \emph{future} portion, $T = T_p + T_f$ (Sec.~\ref{sec:impl}). Let $\phi(\cdot)$ be a transformer encoder and $g(\cdot)$ a lightweight decoder head that maps encoder states back to the feature space.

\subsection{Architecture: encoder-centric transformer}
\begin{figure*}[t]
\centering
\IfFileExists{pathofmarchitecture.png}{
  \includegraphics[width=0.7\linewidth]{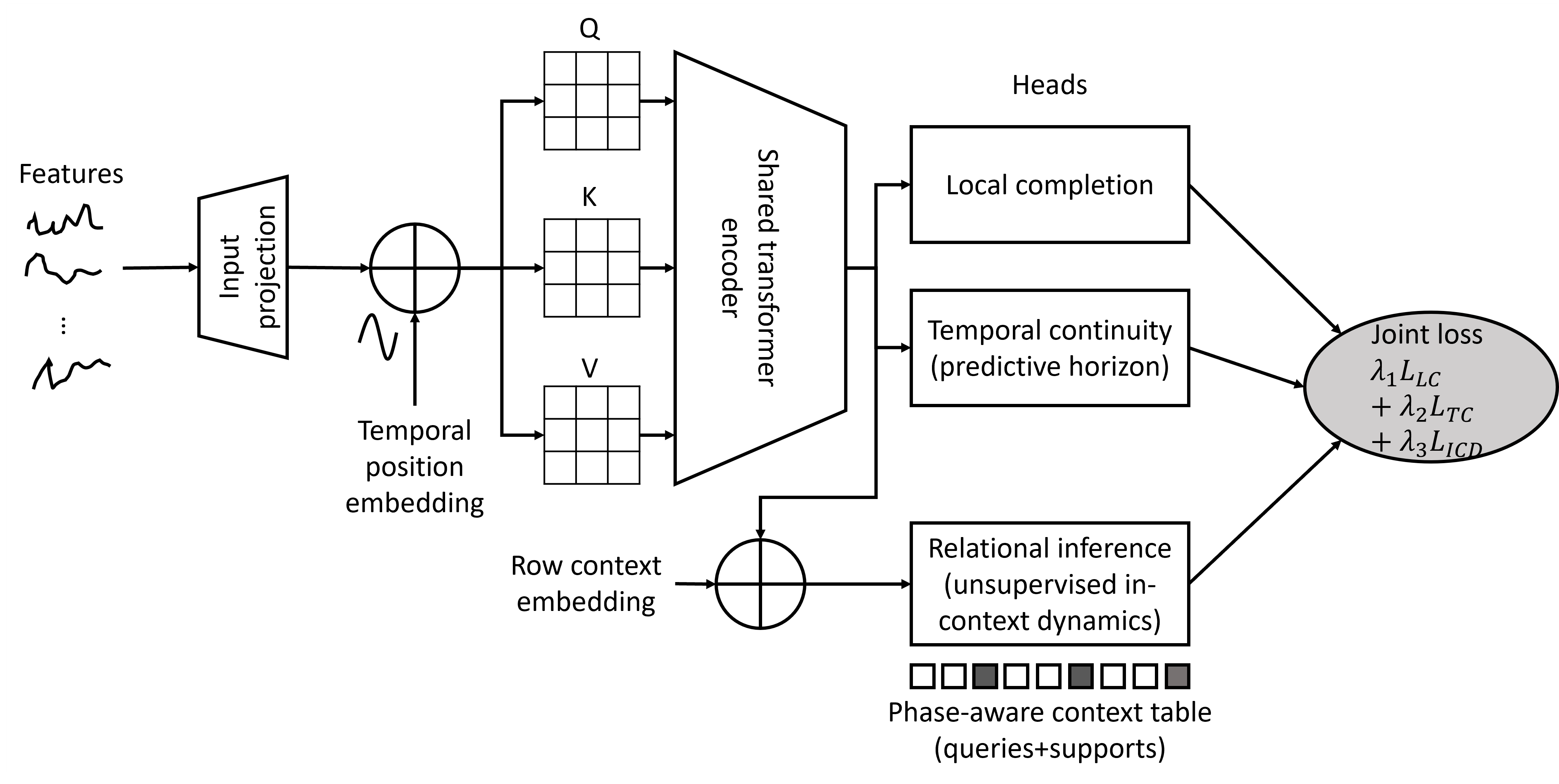}
}{
  \fbox{\parbox{0.95\columnwidth}{\vspace{0.8em}\centering\textbf{Figure placeholder.}\\
  Architecture schematic: encoder-only transformer with three pretraining objectives (LC, TC, uICD).\\
  Provide \texttt{pathofmarchitecture.png} to render this figure.\vspace{0.8em}}}
}
\caption{\textsc{PathoFM} pretraining objectives and architecture. A shared transformer encoder processes multivariate gait windows. (i) \textit{Local Completion} reconstructs masked spans. (ii) \textit{Temporal Continuity} predicts masked futures from observed past. (iii) \textit{Unsupervised In-Context Dynamics} reconstructs a masked query window conditioned on same-subject support windows in a small table, enabling non-parametric adaptation via attention.}
\label{fig:arch}
\end{figure*}

\textsc{PathoFM} (Figure \ref{fig:arch})  is encoder-centric: the transformer encoder bears the burden of representation learning, while each pretext objective uses only a small reconstruction head.

The reference implementation uses:
(i) an input projection $W_{in}:\mathbb{
R}^D \to \mathbb{
R}^{d}$,
(ii) learned or sinusoidal time positional encodings,
(iii) a standard transformer encoder stack, and
(iv) a linear decoder $W_{out}:\mathbb{
R}^{d}\to\mathbb{
R}^{D}$.
For the in-context objective, we additionally embed each \emph{row} of a support--query table with a learned row positional embedding, allowing attention to combine temporal tokens \emph{within} and \emph{across} windows. 

\subsection{Objective 1: Local Completion (LC)}
Local Completion instills a locality/compositionality bias by requiring the model to infer missing contiguous spans from surrounding context.
We sample a boolean mask $\mathbf{M}\in\{0,1\}^{T}$ that covers a set of time indices (typically as a union of contiguous segments) and form a masked input $\widetilde{\mathbf{X}}$ by replacing masked positions with a learned \texttt{[MASK]} token in the embedding space.
The LC loss is mean-squared error on masked positions:
\begin{equation}
\begin{aligned}
\mathcal{L}_{\mathrm{LC}}(\mathbf{X}) 
&= \frac{1}{\sum_t M_t\, D}\sum_{t: M_t=1}\sum_{d=1}^{D}\Big( X_{t,d} - \hat X_{t,d} \Big)^2\\
\quad \hat{\mathbf{X}} &= g(\phi(\widetilde{\mathbf{X}})).
\label{eq:lc}
\end{aligned}
\end{equation}
In the implementation, $\mathbf{M}$ is generated by repeatedly sampling contiguous segments until the desired mask ratio is achieved (Sec.~\ref{sec:impl}).

\subsection{Objective 2: Temporal Continuity (TC)}
Temporal Continuity imposes a dynamics bias by requiring prediction of a mid-horizon continuation from an observed prefix. Let $T_p$ and $T_f$ denote past and future lengths, and let the future index set be $\mathcal{F}=\{T_p+1,\dots,T_p+T_f\}$. We sample a mask $\mathbf{M}^{(f)}$ supported on $\mathcal{F}$ and replace masked future tokens with \texttt{[MASK]}. The TC loss is MSE over the masked future positions:
\begin{equation}
\mathcal{L}_{\mathrm{TC}}(\mathbf{X}) 
= \frac{1}{\sum_{t\in\mathcal{F}} M^{(f)}_t\, D}\sum_{t\in\mathcal{F}:M^{(f)}_t=1}\sum_{d=1}^{D}\Big( X_{t,d} - \hat X_{t,d} \Big)^2.
\label{eq:tc}
\end{equation}
In the strict forecasting setting, we mask \emph{all} future positions ($\mathbf{M}^{(f)}_t=1\ \forall t\in\mathcal{F}$), turning TC into full-horizon continuation.

\subsection{Objective 3: Unsupervised In-Context Dynamics (uICD)}
The uICD objective adds a non-parametric adaptation bias: the model must reconstruct a masked \emph{query} window while attending to a small set of same-subject \emph{support} windows.

\paragraph{Support-query table.}
Given a batch containing multiple windows per subject (enforced by a balanced sampler), we build a small table $\mathcal{T} = [\mathbf{X}^{(1)};\dots;\mathbf{X}^{(R)}]$ of $R$ windows from a single subject. We designate one or more rows as queries and mask either (i) the entire query row or (ii) contiguous spans within the query row.

\paragraph{Row-aware attention.}
We flatten the $R\times T$ temporal tokens into a single sequence and add a learned row embedding to all tokens in a row. Self-attention then allows the query to borrow information from supports without explicit parameter updates.

The uICD objective is used solely during pretraining to shape the representation via attention over same-subject windows. It encourages the encoder to organize latent space such that temporally and structurally related trajectories can be retrieved and composed through attention. \textit{Importantly, our downstream evaluation does not assume access to same-subject support windows at test time: all results are obtained using a frozen encoder and standard probing heads}. While the formulation is compatible with test-time exemplar conditioning when such supports are available (e.g., longitudinal patient monitoring), this is not required for the learned representations to transfer effectively.

The  table-construction choices used for uICD are summarized in Table~\ref{tab:uicdopts}. In the reference configuration, each table contains one fully masked query row, uses same-subject windows as candidate supports, and caps the table size at 16 rows for computational stability. These choices keep the uICD task simple while still allowing the model to learn attention-based, subject-specific reconstruction from exemplar windows.

\paragraph{Loss.}
Let $\mathcal{Q}$ denote query rows and let $\mathbf{M}^{(q)}$ denote a mask on query time steps. The uICD loss is:
\begin{equation}
\begin{aligned}
\mathcal{L}_{\mathrm{uICD}}(\mathcal{T}) 
&= \\ \frac{1}{\sum_{r\in\mathcal{Q}}\sum_t M^{(q)}_{r,t}\, D}&
\sum_{r\in\mathcal{Q}}\sum_{t: M^{(q)}_{r,t}=1}\sum_{d=1}^{D}
\Big( X^{(r)}_{t,d} - \hat X^{(r)}_{t,d} \Big)^2.
\label{eq:uicd}
\end{aligned}
\end{equation}

\subsection{Joint objective}
We pretrain with a weighted sum of objectives:
\begin{equation}
\mathcal{L} = \lambda_{\mathrm{LC}}\,\mathcal{L}_{\mathrm{LC}} + \lambda_{\mathrm{TC}}\,\mathcal{L}_{\mathrm{TC}} + \lambda_{\mathrm{uICD}}\,\mathcal{L}_{\mathrm{uICD}}.
\label{eq:joint}
\end{equation}
In the reference configuration, we set $\lambda_{\mathrm{LC}}=\lambda_{\mathrm{TC}}=\lambda_{\mathrm{uICD}}=1$.

\section{Experimental Setup and Implementation Details}
\label{sec:impl}
\subsection{Data and protocol}
We consider multivariate gait time series from SCI patients (230 subjects), where each trial is represented as a multivariate sequence of gait-related biomechanical variables. Downstream evaluation uses strict \textit{subject holdout}: 10 subjects are held out for test and 10 for validation, with the remainder for training. This protocol evaluates generalization to entirely unseen subjects, reflecting realistic deployment where new patients appear at test time.

\subsection{Downstream tasks and probing protocol}
\label{sec:downstream}
Our goal is \emph{cross-task transfer}: a single pretrained encoder should support both categorical and continuous endpoints with minimal task-specific fine-tuning. We evaluate three classification tasks and one regression task:
(i) \textbf{pathology category} (tetraplegic vs. paraplegic gait patterns),
(ii) \textbf{gender} (binary),
(iii) \textbf{SCIM level \cite{catz1997scim}} (binary; high vs. low functional independence), and
(iv) \textbf{GRF regression} (ground reaction force components).

For each downstream task, we freeze the pretrained encoder $\phi$ and train a lightweight head $h$ on training subjects only (linear probing or a shallow MLP probing depending on task). Classification heads use cross-entropy loss; regression heads use mean-squared error. Early stopping and model selection are performed on the validation-subject split, and we report test performance on the held-out subjects.

\subsection{Windowing and feature selection}
We use biomechanical variables such as angles, progression, power, moment as inputs to the model. Each trial is segmented into fixed-length windows using a sliding window with stride $s$.
We use $T_p=50$ past steps and $T_f=50$ future steps (so $T=100$) and stride $s=10$.

\subsection{Pretraining data composition and augmentations}
The pretraining code 
include both original recordings and additional augmented samples (in our experiments, we use a pre-generated ``augmented'' dataset directory). Conceptually, we favor \textit{dynamics-preserving} transformations: amplitude scaling, mild temporal warping (phase perturbations), and small jitter that do not violate biomechanical plausibility. Such augmentations encourage invariance to noise variation while retaining clinically meaningful phase-locked morphology.

\subsection{Handling missing values via interpolation}
Clinical time series often contain missing values due to sensor dropouts or preprocessing artifacts. Instead of filling with zeros (which introduces an out-of-distribution value for many channels), we apply \textit{per-feature linear interpolation} along time with edge filling: missing values inside a sequence are linearly interpolated between nearest valid neighbors; leading/trailing missing spans are filled with the first/last valid value. 

\subsection{Global min-max normalization}
We compute per-feature global minimum and maximum \emph{across training subjects only}, and normalize all windows to $[0,1]$ using these training statistics. This preserves a consistent scale across subjects and avoids leakage from validation/test subjects.

\subsection{Subject-balanced batching}
\label{sec:balanced}
Clinical datasets are typically imbalanced: some subjects contribute more trials or longer recordings. To prevent the model from being dominated by a few high-volume subjects, we use a \textit{BalancedSubjectBatchSampler} that constructs each batch by sampling $S$ subjects and drawing $W$ windows per subject (with replacement when needed). In our configuration, $S=32$ and $W=4$, giving batch size $B=S\times W=128$. This batching is also crucial for uICD: it ensures each batch contains multiple windows per subject so that support-query tables can be formed.

\subsection{uICD table construction and phase-aware supports}
The uICD objective requires a small support-query table with multiple windows from the same subject. The reference implementation constructs such a table \emph{within each batch}:
\begin{enumerate}[nolistsep]
  \item Group batch indices by subject ID and select a subject with at least two windows in the batch.
  \item From that subject's windows, choose $Q$ query rows (default $Q=1$; multi-query tables are supported) and designate the remaining rows as candidate supports.
  \item Mask the query rows either fully (default) or by masking contiguous spans (controlled by a query mask ratio).
  \item Optionally select \emph{nearby supports}: prioritize windows from the same trial as the query and with closest start index (a phase-aware heuristic), then include additional supports from other trials if needed.
  \item Cap the table size to a maximum number of rows (default 16) for computational stability.
\end{enumerate}

This construction yields a compact, subject-consistent context that allows attention to implement a form of \emph{non-parametric personalization} without labels or gradient updates.

\begin{table}[t]
\centering
\caption{uICD table-construction options.}
\label{tab:uicdopts}
\begin{tabular}{@{}ll@{}}
\toprule
\textbf{Option} & \textbf{Default} \\
\midrule
Queries per table & $Q=1$ \\
Query masking & full-row mask (ratio $=1.0$) \\
Support selection & random \\
Supports per query & all same-subject rows (no $k$-NN cap) \\
Max rows per table & 16 \\
\bottomrule
\end{tabular}
\end{table}

\subsection{Masking scheme}
Both LC and TC use contiguous-span masking. To generate a mask with target coverage $\rho$, we repeatedly sample segment lengths $\ell\sim \mathrm{Unif}(\ell_{min},\ell_{max})$ and start indices uniformly in the allowed range until the mask covers approximately $\rho$ fraction of the region.

In the reference configuration: LC masks $\rho_{\mathrm{LC}}=0.8$ of the window using segment lengths $\ell\in[4,16]$, while TC masks $\rho_{\mathrm{TC}}=1.0$ of the future region using automatically chosen segment lengths proportional to the horizon (here $\ell\in[5,25]$).

\subsection{Model and optimization}
\label{sec:opt}
We use a transformer encoder with embedding dimension $d=128$, depth of eight, four attention heads, dropout of 0.1, and feed-forward width multiplier of two. The encoder is pre-normalized. We optimize with AdamW (learning rate $10^{-4}$, weight decay $10^{-3}$), apply gradient clipping at 1.0, and use cosine annealing over 200 epochs with early stopping based on validation pretext loss. Table~\ref{tab:hparams} consolidates the reference implementation hyperparameters, including the windowing scheme, batching strategy, masking ratios, transformer configuration, optimizer, and training schedule. Unless otherwise stated, all reported pretraining and downstream evaluations use these settings.

\begin{table}[t]
\centering
\caption{Reference implementation hyperparameters}
\label{tab:hparams}
\resizebox{0.8\textwidth}{!}{
\begin{tabular}{@{}ll@{}}
\toprule
\textbf{Category} & \textbf{Setting} \\
\midrule
Window lengths & $T_p=50$, $T_f=50$ (total $T=100$); stride $s=10$ \\
Batching & $S=32$ subjects/batch, $W=4$ windows/subject (batch $B=128$) \\
Objectives & $\lambda_{\mathrm{LC}}=\lambda_{\mathrm{TC}}=\lambda_{\mathrm{uICD}}=1$ \\
LC masking & mask ratio 0.80; segment length $[4,16]$ \\
TC masking & future mask ratio 1.00; segment length $[5,25]$ \\
Transformer & $d=128$, depth 8, heads 4, dropout 0.1, FF mult 2 \\
Optimizer & AdamW, LR $10^{-4}$, weight decay $10^{-3}$ \\
Schedule & cosine annealing, 200 epochs, early stop patience 20 \\
\bottomrule
\end{tabular}}
\end{table}

\section{Results and Analysis}
\label{sec:results}
We evaluate (i) \textit{pretext generalization} under strict subject holdout for Local Completion (LC) and Temporal Continuity (TC), and (ii) \textit{downstream transfer} to heterogeneous endpoints: classification and continuous regression using frozen pretrained encoders with lightweight probe heads. Strict subject holdout reflects the clinically relevant regime where the test set contains entirely unseen subjects.

\subsection{Pretext generalization under subject holdout}

For LC and TC, we measure $R^2$ and Pearson correlation $r$ between reconstructed predictions and ground truth on held-out subjects (evaluated only on masked positions).
Table~\ref{tab:pretext} shows that both objectives generalize strongly to unseen subjects, suggesting that the learned representations encode fundamental gait morphology and dynamics rather than memorizing subject identity. 

\begin{figure*}[t]
\centering
\IfFileExists{gaitoverlay.png}{
  \includegraphics[width=0.75\linewidth]{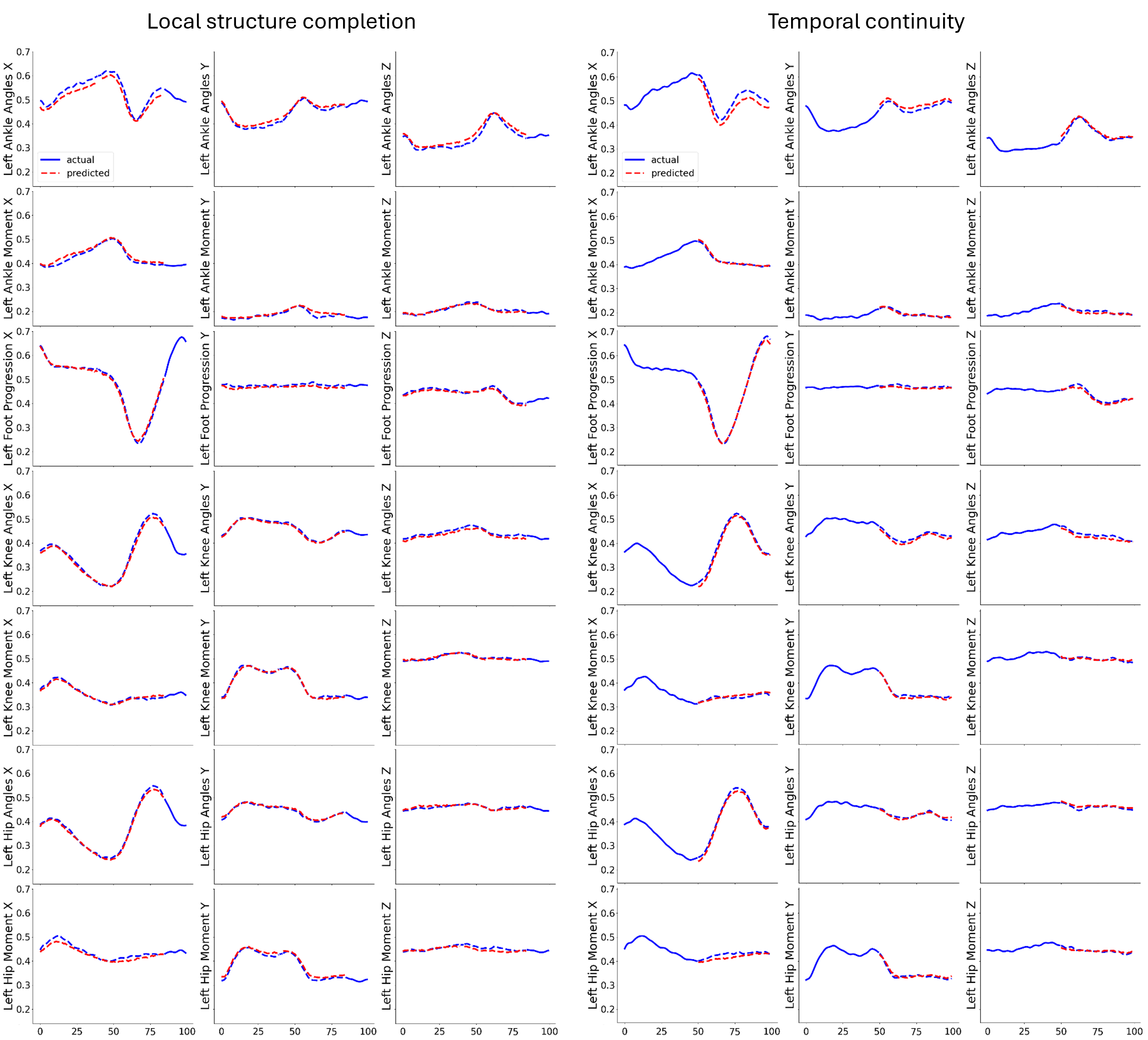}
}{
  \fbox{\parbox{0.95\columnwidth}{\vspace{0.8em}\centering\textbf{Figure placeholder.}\\
  Qualitative overlay of ground-truth vs. predicted gait trajectories for masked completion and continuation on an unseen subject.\\
  Provide \texttt{gaitoverlay.png} to render this figure.\vspace{0.8em}}}
}
\caption{Qualitative pretext sanity check on an unseen SCI subject. Ground truth (solid) vs. \textsc{PathoFM} predictions (dashed) across multiple joint angles and ground reaction force (GRF) components over a normalized gait window.}
\label{fig:overlay}
\end{figure*}

The qualitative reconstructions in Figure~\ref{fig:overlay} provide a complementary sanity check on these quantitative results. Across joint-angle and GRF channels, the predicted trajectories closely follow the ground-truth curves on an unseen SCI subject, suggesting that the pretrained encoder captures both phase-aligned waveform morphology and channel-specific temporal dynamics.

\begin{table}[t]
\centering
\caption{Pretext task performance on held-out subjects. Higher is better.}
\label{tab:pretext}
\begin{tabular}{@{}lcc@{}}
\toprule
\textbf{Pretext task} & $\mathbf{R^2}$ $\uparrow$ & \textbf{Pearson $r$} $\uparrow$ \\
\midrule
Local Completion (LC) & 0.90 & 0.95 \\
Temporal Continuity (TC) & 0.85 & 0.92 \\
\bottomrule
\end{tabular}
\end{table}

\subsection{Downstream transfer: ablations}

We ablate the three-objective mixture by removing one component at a time and evaluate transfer to three classification tasks (pathology category, gender, SCIM independence level) and GRF regression.
Table~\ref{tab:ablations} shows that the full mixture is the most balanced overall: removing any component degrades at least one task family. Adding uICD substantially improves GRF metrics across axes. This is consistent with uICD behaving like non-parametric personalization: query reconstruction can borrow cues consistent across subjects from supports, which is valuable for continuous targets. By contrast, adding exemplar conditioning using uICD does not add useful information for gender classification. In summary, the multi-bias recipe that we used avoids a single-task optimum in exchange for a consistent cross-task competence under subject shift.

\begin{table*}[t]
\centering
\small
\resizebox{\textwidth}{!}{%
\begin{tabular}{
l*{12}{c}
}
\toprule
& \multicolumn{2}{c}{Pathology category} & \multicolumn{2}{c}{Gender} & \multicolumn{2}{c}{SCIM} & \multicolumn{2}{c}{GRF\_X} & \multicolumn{2}{c}{GRF\_Y} & \multicolumn{2}{c}{GRF\_Z} \\
\cmidrule(lr){2-3}\cmidrule(lr){4-5}\cmidrule(lr){6-7}\cmidrule(lr){8-9}\cmidrule(lr){10-11}\cmidrule(lr){12-13}
Method & {F1 ↑} & {AUC ↑} & {F1 ↑} & {AUC ↑} & {F1 ↑} & {AUC ↑} & {$\rho$ ↑} & {RMSE} & {$\rho$} & {RMSE ↓} & {$\rho$ ↑} & {RMSE ↓} \\
\midrule
TC+uICD     & 0.68 & 0.64 & 0.762 & 0.54 & 0.70 & 0.64 & 0.65 & 0.021 & 0.83 & 0.033 & \textbf{0.89} & \textbf{0.199} \\
LC+uICD    & 0.69 & 0.64 & 0.770 & 0.54 & 0.70 & 0.64 & 0.70 & 0.020 & 0.83 & 0.033 & 0.89 & 0.200 \\
LC+TC     & 0.66 & 0.64 & \textbf{0.780} & \textbf{0.61} & 0.69 & 0.60 & 0.64 & 0.022 & 0.81 & 0.035 & 0.84 & 0.246 \\
\midrule
\textbf{LC+TC+uICD} & \textbf{0.69} & \textbf{0.66} & 0.764 & 0.55 & \textbf{0.71} & \textbf{0.65} & \textbf{0.71} & \textbf{0.020} & \textbf{0.83} & \textbf{0.033} & 0.88 & 0.202 \\
\bottomrule
\end{tabular}%
}
\caption{Performance across downstream classification and prediction tasks when each component of the loss term is ablated. Arrows indicate whether higher (↑) or lower (↓) values are better. The full loss term performs best in four out of six tasks, illustrating the importance of the combined loss in learning generalizable representations. }
\label{tab:ablations}
\end{table*}

\paragraph{Interpretation.}
LC preserves local morphology that aids classification; TC and uICD are especially beneficial for continuous outcomes where phase-consistent dynamics and magnitude fidelity matter. The full mixture benefits from the complementarity of these biases: LC teaches \emph{what} local patterns look like, TC teaches \emph{how} they evolve, and uICD teaches \emph{when} to adapt from exemplars.

\subsection{Benchmarks by inductive-bias family}
We compare \textsc{PathoFM} to representatives from grouping-based pretraining (contrastive, DINO-style prototypes, subject-ID), generative diffusion, and a diffusion+reconstruction hybrid. Table~\ref{tab:benchmarks} shows a recurring pattern: grouping objectives are competitive for classification but underperform on regression, while generative objectives improve continuous targets but may not maximize class separation. \textsc{PathoFM} balances both.

\begin{table*}[!h]
\centering
\small
\resizebox{0.75\textwidth}{!}{%
\begin{tabular}{l *{8}{c}}
\toprule
& \multicolumn{2}{c}{Pathology category} & \multicolumn{2}{c}{GRF\_X} & \multicolumn{2}{c}{GRF\_Y} & \multicolumn{2}{c}{GRF\_Z} \\
\cmidrule(lr){2-3}\cmidrule(lr){4-5}\cmidrule(lr){6-7}\cmidrule(lr){8-9}
Method & F1 ↑ & AUC ↑ & $\rho$ ↑ & MSE ↓ & $\rho$ ↑ & MSE ↓ & $\rho$ ↑ & MSE ↓ \\
\midrule
Dino               & 0.67 & 0.62 & 0.67 & 0.020 & 0.82 & 0.035 & \textbf{0.89} & 0.230 \\
Contrastive        & \underline{0.68} & \textbf{0.68} & 0.63 & 0.021 & 0.78 & 0.038 & 0.78 & 0.281 \\
subject identification           & 0.63 & 0.65 & 0.69 & 0.020 & 0.82 & 0.034 & 0.86 & 0.224 \\
Diffusion only     & 0.63 & 0.65 & 0.58 & 0.024 & 0.73 & 0.043 & 0.84 & 0.230 \\
Diffusion+LC+uICD  & 0.61 & 0.64 & \underline{0.70} & \textbf{0.019} & \textbf{0.83} & \textbf{0.033} & 0.87 & \underline{0.212} \\
\midrule
\textbf{LC+TC+uICD}         & \textbf{0.69} & \underline{0.66} & \textbf{0.71} & \underline{0.020} & \textbf{0.83} & \textbf{0.033} & \underline{0.88} & \textbf{0.202} \\
\bottomrule
\end{tabular}%
}
\caption{Benchmark performance comparison by inductive bias family across pathology category classification and GRF prediction tasks. Arrows indicate whether higher (↑) or lower (↓) values are better. The combined loss term that we propose performs best in one or both the metrics reported for each task. }
\label{tab:benchmarks}
\end{table*}

\section{Discussion: What Biases Help, When, and Why?}
The results highlight that inductive bias is not an abstract philosophical garnish; it is an operational design axis with predictable trade-offs.
\vspace{-0.4cm}
\paragraph{Grouping bias: strong discrimination, weaker magnitude fidelity.}
Grouping-based objectives (contrastive, prototypes, subject-ID) produce representations that separate instances or subjects, which benefits classification. But they can discard calibrated magnitude information and fine-grained waveform details, exactly what continuous kinetic tasks require.
\vspace{-0.3cm}
\paragraph{Dynamics bias: essential for continuous endpoints.}
Temporal continuity (forecasting/continuation) directly trains the representation to encode phase-consistent trajectories. This aligns with kinetic regression tasks, where the model must preserve timing and rate-of-change.
\vspace{-0.3cm}
\paragraph{Reconstruction bias: preserves local structure, but needs a discriminative complement.}
Masked reconstruction improves signal fidelity and is naturally aligned with denoising and imputation. However, if used alone it may not explicitly enlarge class margins.
\vspace{-0.3cm}
\paragraph{In-context bias: non-parametric personalization.}
uICD acts like a learned ``compare to similar examples'' procedure. It can improve cross-subject generalization when test-time supports from the same subject (or clinic) are available. The main practical constraint is support selection: irrelevant supports can mislead attention.
\vspace{-0.3cm}
\paragraph{A mechanistic view: uICD as attention-based exemplar regression.}
It is useful to interpret uICD through a simple lens: in a transformer, a token representation is updated by attention-weighted combinations of value vectors. If we denote (for a single head) the query/key/value matrices by $Q,K,V$ and the attention output by
\begin{equation}
\mathrm{Attn}(Q,K,V) = \mathrm{softmax}\!\left(\frac{QK^\top}{\sqrt{d_k}}\right)V,
\end{equation}
then a masked query token can be viewed as performing a learned similarity search over support tokens (via $QK^\top$) and assembling an estimate from retrieved values. In uICD, row embeddings encourage the model to treat windows as distinct exemplars while still allowing cross-row retrieval. Under this view, ``phase-aware'' support selection (same trial, nearby start index) is not merely a heuristic: it increases the probability that the nearest neighbors in representation space correspond to biomechanically aligned events (e.g., heel strike, toe-off), which in turn stabilizes attention-based reconstruction.
\vspace{-0.3cm}

\paragraph{Why balancing and interpolation matter in clinical time series.}
Two implementation choices from Sec.~\ref{sec:impl} deserve emphasis because they directly shape what the model can learn. (i) Subject-balanced batching reduces spurious shortcuts: without it, the encoder can partially ``solve'' uICD by memorizing high-frequency subjects. Balancing forces the model to explain within-subject structure rather than exploit data volume. (ii) Linear interpolation with edge fill treats missingness as a smooth measurement artifact rather than a special symbol. This is appropriate for gait kinematics/kinetics, where abrupt zeros are biomechanically implausible and can become a misleading cue.
\vspace{-0.3cm}
\paragraph{Design guidance.}
For clinical foundation models with limited data and compute:
(i) if regression and calibration matter, include a dynamics objective (forecasting/continuation) and consider in-context conditioning;
(ii) if classification dominates, include local reconstruction and optionally a light grouping objective (but avoid letting grouping dominate);
(iii) for balanced transfer, combine LC+TC+uICD and tune weights.
\vspace{-0.2cm}
\vspace{-0.2cm}
\section{Conclusion}
We analyzed time-series pretraining through the lens of inductive bias using pathological gait as a case study. We showed that a dynamics-centric mixture of Local Completion, Temporal Continuity, and Unsupervised In-Context Dynamics yields balanced transfer to both classification and regression under strict subject holdout. Beyond this specific domain, the larger lesson is that pretraining objectives are not interchangeable: each encodes assumptions about what structure matters, and careful combinations can achieve robustness that single-bias objectives struggle to match.

\section*{Funding declaration}
This study was partially funded by the Schweizer Paraplegiker Stiftung (2021-HS-348), within the Digital Transformation in Personalised Healthcare initiative for individuals with spinal cord injury, and the Innosuisse innovation project number: 115.290 IP-ICT.

\bibliography{references}
\bibliographystyle{unsrt}

\end{document}